\documentclass[10pt,twocolumn,a4paper,conference]{IEEEtran}
\usepackage[utf8]{inputenc}
\usepackage{tikz}
\usetikzlibrary{positioning}
\usetikzlibrary{calc}
\usepackage{amsmath}
\usepackage{amssymb}
\usepackage{amsfonts}
\usepackage{setspace}
\usepackage{graphicx}
\usepackage{caption}
\usepackage[T1]{fontenc}
\usepackage[utf8]{inputenc}
\usepackage{epstopdf}
\usepackage{xcolor}
\usepackage{multirow}
\usepackage{hyperref}
\usepackage[acronym,shortcuts]{glossaries}
\makenoidxglossaries
\usepackage[capitalize]{cleveref}
\usepackage{bm}
\usepackage{algorithmic}
\usepackage[ruled]{algorithm}
\usepackage{microtype}
\usepackage{cite}

\newcommand{\mbf}{\mathbf}
\newcommand{\mbfh}[1]{\mathbf{#1}^{\mathsf{H}}}

\newacronym{AP}{AP}{Access Points}
\newacronym{CSI}{CSI}{channel state information}
\newacronym{ECDF}{ECDF}{empirical cumulative distribution function}
\newacronym{MIMO}{MIMO}{multiple-input multiple-output}
\newacronym{MISO}{MISO}{multiple-input single-output}
\newacronym{OFDM}{OFDM}{orthogonal frequency division multiplexing}
\newacronym{SINR}{SINR}{signal to interference and noise ratio}
\newacronym{SNR}{SNR}{signal to noise ratio}
\newacronym{SLNR}{SLNR}{signal to leakage and noise ratio}
\newacronym{wSLNR}{wSLNR}{weighted SLNR}
\newacronym{DNN}{DNN}{Deep neural network}
\newacronym{QoS}{QoS}{Quality of Service}
\newacronym{cfmimo}{CFMIMO}{cell-free multiple-input multiple-output}
\newacronym{TDD}{TDD}{time-division duplexing}
\newacronym{UL}{UL}{uplink}
\newacronym{DL}{DL}{downlink}
\newacronym{WiT}{WiT}{wireless transformer}
\newacronym{BS}{BS}{base station}
\newacronym{UE}{UE}{user equipment}
\newacronym{MGD}{MGD}{multi-gradient decent}
\newacronym{MHA}{MHA}{multi-head attention}
\newacronym{FFN}{FFN}{feed-forward network}
\newacronym{FC}{FC}{fully connected}
\newacronym{MLP}{MLP}{multi layer perceptron}
\newacronym{MRT}{MRT}{maximum ratio transmission}
\newacronym{ZF}{ZF}{zero forcing}
\newacronym{SR}{SR}{sum rate}

\allowdisplaybreaks

\begin{document}



\title{DNN-Enabled Multi-User Beamforming for Throughput Maximization under Adjustable Fairness}
\author{Kaifeng Lu, Markus Rupp, Stefan Schwarz\\
Institute of Telecommunications, Technische Universität (TU) Wien \\
Email: kaifeng.lu@tuwien.ac.at
}
\maketitle
\begin{minipage}{500pt}
\vspace{-350pt}
\centering{
\footnotesize{© 2026 IEEE. Personal use of this material is permitted. This is the author's version of the work accepted for publication in IEEE International Conference on Machine Learning in Communications and Networking (ICMLCN 2026).}}
\end{minipage}

\begin{abstract}  
 Ensuring user fairness in wireless communications is a fundamental challenge, as balancing the trade-off between fairness and sum rate leads to a non-convex, multi-objective optimization whose complexity grows with network scale. To alleviate this conflict, we propose an optimization-based unsupervised learning approach based on the wireless transformer (WiT) architecture that learns from channel state information (CSI) features. We reformulate the trade-off by combining the sum rate and fairness objectives through a Lagrangian multiplier, which is updated automatically via a dual-ascent algorithm. This mechanism allows for a controllable fairness constraint while simultaneously maximizing the sum rate, effectively realizing a trace on the Pareto front between two conflicting objectives. Our findings show that the proposed approach offers a flexible solution for managing the trade-off optimization under prescribed fairness.
\end{abstract} 

\glsresetall

\section{Introduction}
\label{sec:Intro}
\acp{DNN} have shown strong potential for addressing a wide range of problems in next-generation wireless networks. As the network scales, \ac{CSI} is considered as one of the essential learning features, providing a rich, unified representation that captures key spatiotemporal relationships and propagation characteristics across numerous tasks. Leveraging \ac{CSI} as the input, \acp{DNN} have been applied to solve multiple problems such as radio resource management and joint communication-and-sensing, yielding robust, low-latency decisions in interference-limited, time-varying channels. In particular, \acp{DNN} have been used to tackle localization~\cite{10382964,9348191,9833994}, user clustering in cell-free MIMO~\cite{9833939}, channel prediction~\cite{8752012,11152870}, and energy-efficiency aware power control for green communications~\cite{10498103,9136914,9841466} among other tasks.Thus, \ac{CSI} serves as a bridge between machine learning and wireless communication tasks, enabling consistent and efficient solutions.

As mobile systems keeps evolving, the \ac{QoS} requirements are becoming more strict and diverse, aiming to deliver comparable benefits for all users~\cite{10198239}. However, ensuring users' fairness across the network remains a fundamental challenge, as the feasible fairness solutions are characterized by the Pareto front in a non-convex multi-objective sum rate trade-off \cite{yeung2023biobjective}. To address this, \acp{DNN} have emerged as a practical alternative to complex \ac{MGD} algorithms. In~\cite{9991855}, a weighted-sum objective with deep reinforcement learning tackles the non-convex optimization, whereas~\cite{10994230} uses a transformer-based DNN, training for power allocation that maximizes the minimum spectral efficiency across subcarriers. A common limitation is that using fixed weights or a max–min formulation cannot guarantee a certain level of fairness. In practice, fairness tends to decrease when maximizing the sum rate, making the achieved fairness highly dependent on the weight selection.


To formulate a solution that maximizes throughput while achieving a prescribed fairness level, we propose an unsupervised learning approach based on the transformer architecture in which the fairness target is explicitly identified. 
\begin{itemize}
    \item We formulate a composite objective that couples the sum rate with a fairness constraint measured by Jain’s index $J$, which is not the only possible fairness definition but is well suited to our setting; see \cref{sec2A} for details. Unlike ~\cite{9991855}, which uses a convex combination, we adopt a Lagrangian approach with a hinge penalty on the sum rate that only activates when a violation occurs on the target fairness lower bound $J_{LB}$.
    \item  A controllable dual multiplier $\lambda$ scales the penalty, and an adaptive dual-ascent scheme automatically updates $\lambda$ to ensure the fairness constraint is satisfied. Then the sum rate is maximized at the target fairness.
    \item By selecting a set of lower bound values $J_{LB}$ for DNN training, we obtain separate points in the sum rate versus fairness plane whose trace approximates the Pareto front, thereby enabling us to specify the maximized sum rate at prescribed fairness levels.
\end{itemize} 


\section{System Model and Problem Formulation}
\label{sec:System}
We consider a multi-user \ac{MIMO} downlink broadcast transmission with a single transmitter base station equipped with $N_t$ antennas, serving $N_u$ single antenna \acp{UE}. Each broadcast link is a frequency-flat narrow-band \ac{MISO} channel where we consider a single subcarrier of an \ac{OFDM} transmission.
\subsection{Downlink Input-Output Relationship}
\label{sec2A}
For each user $u \in \mathcal{U} \triangleq \{1,\dots,N_u\}$, the channel vector is represented by $\mbf{h}_u \in\mathbb{C}^{N_t \times 1}$. Let $\mbf{f}_u \in \mathbb{C}^{N_t \times 1}$ denote the beamforming vector of user $u$, and let $x_u$ denote user $u$'s transmit signal. The overall received signal at user $u$ is:
\begin{gather}
    y_u = \mathbf{h}_u^{\mathrm{H}}\mathbf{f}_u x_u +
    \mathbf{h}_u^{\mathrm{H}} \sum_{\substack{\ell \in \mathcal{U} \setminus \{u\}}} \mathbf{f}_\ell x_\ell + n_u,
\end{gather}
where, $n_u \sim \mathcal{CN}(0, \sigma_u^2)$ is complex Gaussian noise.

In the input-output relationship, $\mathbf{h}_u^{\mathrm{H}}\mathbf{f}_u x_u$ is the useful signal at target user $u$, whereas the summation term represents the inter-user interference from other users $\ell \neq u$, and vector $\mathbf{f}_\ell$ denotes the beamformer of the $\ell$-th interfering user. We assume $\mathbb{E}\!\left(|x_u|^2\right) = 1$, a total power budget at the \ac{BS} of $P_{tot}$ with equal power allocation to all users, and the subcarrier bandwidth is $B$. Specifically, we can decompose the beamformer of any user $u \in\ \mathcal{U}  $ as:
\begin{gather}
    \mathbf{f}_{u} = P_u \tilde{\mathbf{f}}_{u},\quad\|\tilde{\mathbf{f}}_{u}\|^2 = 1,\quad P_u =\sqrt{\frac{P_{tot}}{N_u}.}
\end{gather}

With the abovementioned assumptions, the instantaneous \ac{SINR} and the achievable throughput of user $u$ are: 
\begin{gather}
    \mathrm{SINR}_u = \frac{|\mathbf{h}_u^{\mathrm{H}}\mathbf{f}_u|^2}
       {\displaystyle \sum_{\substack{\ell=1, \ell \neq u}}^{N_u}
        |\mathbf{h}_u^{\mathrm{H}}\mathbf{f}_\ell|^2 + \sigma_u^2 } \\
    R_u = \log_2\!\left(1 + \mathrm{SINR}_u\right).
\end{gather}
\paragraph{Fairness Index} The system's fairness level is quantified by Jain's index, calculated from each user's achievable throughput~\cite{5963489}:
\begin{gather}
\label{jain's index}
    J = \frac{\left(\sum_{u = 1}^{N_u} R_u \right)^2}{N_u \sum_{u = 1}^{N_u} R_u^2}.
\end{gather}
Jain’s fairness index is a function of the user rate $R_u$; it lies in $(1/N_u,1)$ and increases as the user rates become more equal. Consequently, it is coupled closely with the sum-rate through the same variable $R_u$, resulting in a composite objective that is consistent with $R_u$ and exhibits stable gradients during training.

\paragraph{Weighted SLNR Beamforming}
Following our prior work \cite{11202811}, where we adopt a \ac{wSLNR} approach, we use this method as a heuristic benchmark in our simulations and as a clear, interpretable baseline for comparison with our DNN-based beamforming:
\begin{gather}
\label{eq:wSLNR}
    \text{wSLNR}_{u} = \frac{\left| \mbfh{h}_{u} \mathbf{f}_{u} \right|^2}{\sum_{\ell = 1,\ell\neq u}^{N_u} \omega_{\ell} \left| \mbfh{h}_{\ell} \mathbf{f}_{u} \right|^2 + \sigma_n^2}.
\end{gather}
Here, $\left| \mbfh{h}_{\ell} \mathbf{f}_{u} \right|^2$ denotes the leakage power from user $u$ to other user $\ell$. The weight $\omega_\ell$ serves as a penalty factor that prioritizes users suffering from a weak channel gain, while penalizing those with stronger channels. The weights $\omega_\ell$ are calculated by taking the inverse of the channel gain and raising it to a tunable exponent $\alpha_\ell$:
\begin{gather}
\label{eq:weighted_value}
    \omega_{\ell } = \frac{\tilde{\omega}_{\ell }}{\sum_{\ell \in \mathcal{U}} \tilde{\omega}_{\ell}}, \quad
    \tilde{\omega}_{\ell } = \frac{1}{\left( \left\| \mathbf{h}_{\ell } \right\|^2 \right)^{\alpha_\ell}}.
\end{gather}
Finally, the optimal beamforming vector can be derived from the Rayleigh quotient:
\begin{gather}
\label{eq:wbeamformer}
    \mbf{f}_{u} \propto \left( \sum_{\ell = 1,\ell\neq u}^{N_u} \omega_{\ell} \mbf{h}_{\ell} \mbfh{h}_{\ell} + \sigma_n^2 \mathbf{I}_{N_t} \right)^{-1} \mbf{h}_{u}.
\end{gather}

\subsection{Problem Formulation}
Our target is to develop a general and flexible method for maximizing the downlink sum throughput while maintaining fairness across users. In other words, for a given Jain’s index lower bound $J_{LB}$, we aim to find the Pareto optimal sum rate.
\begin{equation}
\label{eq:sumrate_fair_opt}
\begin{aligned}
\max_{\{\mathbf{f}_u\}_{u=1}^{N_u}} \quad & \sum_{u=1}^{N_u} R_u \\
\text{s.t.}\quad 
& J \ge J_{LB}, \\
& \sum_{u=1}^{N_u} \|\mathbf{f}_u\|^2 \le P_{tot} .
\end{aligned}
\end{equation}
Here, we isolate the contribution of the power domain, avoiding joint impact on balancing the fairness, so that we can obtain a deployment-agnostic solution that remains applicable when power control is unavailable. 
\section{Transformer-Based Sum Rate Fairness Optimization}
\label{sec:WiT}
\begin{figure*}[t] 
  \begingroup
  \setlength{\textfloatsep}{6pt}      
  \setlength{\abovecaptionskip}{2pt}  
  \setlength{\belowcaptionskip}{-2pt}
  \centering
  \includegraphics[width=\textwidth, height=0.70\textheight, keepaspectratio]{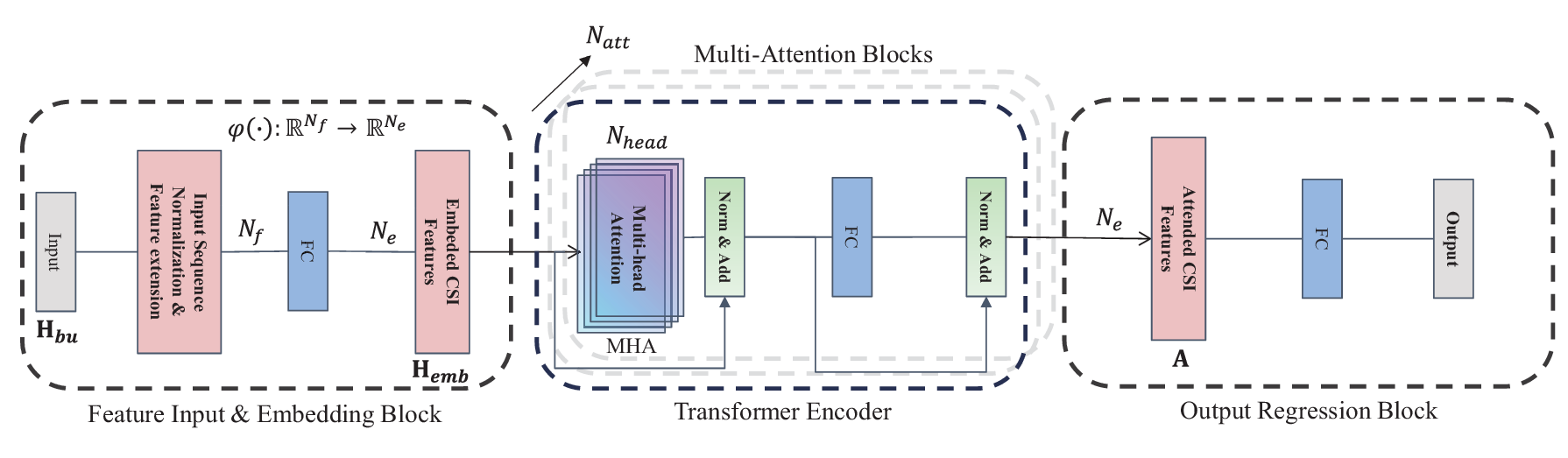}
  \caption{Transformer-DNN architecture for throughput maximization under adjustable fairness}
  \label{fig:WiT}
  \endgroup
\end{figure*}
The transformer \ac{DNN} architecture was first introduced in~\cite{vaswani2023attentionneed} and subsequently extended to the wireless domain with a task-specific \ac{WiT} in \cite{9833994}. While the prior study used \ac{CSI}-based learning for resource management and localization~\cite{9833994}, here we broaden the scope to the fairness-oriented beamforming task.

\subsection{Training Data}
 We take two sub-steps before inputting the \ac{CSI} sequence into the WiT. We begin by decomposing the \ac{CSI} into a normalized channel vector and \acp{SNR}, which makes a numerically stable representation that effectively handles the large dynamic range and variance of wireless signals while reducing the gradient fluctuation for robust training: 
\begin{gather}
\tilde{\mathbf{h}}_{u}=\frac{\mathbf{h}_{u}}{\lVert \mathbf{h}_{u}\rVert},\qquad
\beta_{u}=\frac{\lVert \mathbf{h}_{u}\rVert^{2}}{\sigma_{u}^{2}}.
\end{gather}
Then, we split the normalized channel vector into real and imaginary parts, and take the logarithms of the \ac{SNR}. For each user, the concatenated \ac{CSI} feature vector is: $ \mathbf{h}=\big[\,\Re(\tilde{\mathbf{h}}_{u}),\ \Im(\tilde{\mathbf{h}}_{u}),\ 10\log_{10}(\beta_{u})\,\big]$. The feature length is $N_f = 2N_t +1$. By vertically stacking these features for all users $u\in \mathcal{U}$, we form the \ac{CSI} input sequence $\mathbf{H} \in \mathbb{R}^{N_u \times N_f}$, where each row represents each user's \ac{CSI} feature.
Let the complete dataset be $\mathcal{H}_{\mathrm{tot}}=\big\{\mathbf{H}_n\big\}_{n=1}^{N_{\mathrm{tot}}}\quad
\mathbf{H}_n\in\mathbb{R}^{N_{\mathrm{u}}\times N_{\mathrm{f}}}$, with the total number of data $
\big|\mathcal{H}_{\mathrm{tot}}\big|=N_{\mathrm{tot}}$. We partition it into three disjoint subsets: training set $\mathcal{H}_{\mathrm{train}}$, validation set $\mathcal{H}_{\mathrm{val}}$, and test set $\mathcal{H}_{\mathrm{test}}$, where $\mathcal{H}_{\mathrm{tot}}
\triangleq \mathcal{H}_{\mathrm{train}} \,\cup\, \mathcal{H}_{\mathrm{val}} \,\cup\, \mathcal{H}_{\mathrm{test}}$, with the number of sequences $\big|\mathcal{H}_\mathrm{train}\big|=N_{\mathrm{train}},\quad   \big|\mathcal{H}_{\mathrm{val}}\big|=N_{\mathrm{val}}, \quad \big|\mathcal{H}_{\mathrm{test}}\big|=N_{\mathrm{test}}$ respectively.

\subsection{Model Architecture}
Building on the foundation in \cite{9833994},  ~\cref{fig:WiT} demonstrates a transformer-DNN architecture adopted in this work. It balances fairness while optimizing the overall sum rate by using an explicit loss function.
\paragraph{Feature Input and Embedding Block} For each input batch of size $N_{b}$, we first normalize the random batch selection from the training set. Then we sample (without repetition) an index set $\mathbb{I}_b\subseteq\{1,\dots,N_{\mathrm{train}}\},\quad
\lvert\mathbb{I}_b\rvert=N_b$ and stack the selected sequences to form the batch tensor: $
\mathbf{H}_{bu}=\mathrm{Stack}\big(\{\mathbf{H}_n: n\in\mathbb{I}_b\}\big)\in\mathbb{R}^{N_b \times N_u \times N_f}$.
We embed the $\mathbf{H}_{bu}$ features into an $N_e$-dimensional space using a \ac{FC} layer $\varphi:\mathbb{R}^{N_f}\to\mathbb{R}^{N_e}$.
The embedded features are $\mathbf{H}_{emb}=\varphi(\mathbf{H}_{bu})\in\mathbb{R}^{N_b\times N_u\times N_e}$, which are then fed into the transformer encoder.
\paragraph{Transformer Encoder} The transformer encoder consisting of $N_{att}$ multi-attention blocks with $N_{head}$ parallel \ac{MHA} layers and a \ac{FC} layer ~\cite{vaswani2023attentionneed,LIN2022111}. In each transformer encoder, the input and output shapes are identical so that the attended \ac{CSI} features at the output are $\mathbf A\in\mathbb{R}^{N_b\times N_u\times N_e}$.

Our task primarily requires the intra-sequence dependencies; therefore, $N_{head}$ is a key hyperparameter, since multiple heads allow each head to focus on different relationships among the \ac{CSI} sequence elements.
\paragraph{Output Regression Block} At the output block, we employ a \ac{FC} layer for regression design. Since the model must infer continuous-valued targets from \ac{CSI} features, a regression over a \ac{FC} layer is sufficient.
In our task, we first slice the attended tensor $\mathbf{A}$ along the user's dimension over each batch, obtaining $\mathbf{a}_u \triangleq \mathbf{A}(:,u,:) \in \mathbb{R}^{N_b \times N_e},\quad u=1,\ldots,N_u$. Each $\mathbf{a}_u$ is then sent through the \ac{FC} layer, which produces a regression output --- a quasi-beamformer for each user $\tilde{\mathbf{f}}_u \in \mathbb{R}^{2Nt \times N_b}$ formed by horizontally stacking the real parts and imaginary parts. The predicted beamformer of the \ac{FC} layer's output is normalized by $\hat{\mathbf{f}}_{u}(:,k) =\tilde{\mathbf{f}}_u(:,k) \big/ \lVert \tilde{\mathbf{f}}_u(:,k) \rVert $, where the normalization is applied on each column. The predicted beamformers of all users are then used to calculate the training loss.
\subsection{Loss Function and Training Algorithm}
Our major goal is to solve the maximization problem in \cref{eq:sumrate_fair_opt}, but two practical issues arise. First, maximizing the sum rates $\sum_{u=1}^{Nu} R_u$ naturally pushes down the Jain index $J$ in \cref{jain's index}. Furthermore, $J$ varies during training, so the hard constraint \(J \ge J_{LB}\) cannot be satisfied at every iteration. Second, the DNN can naturally optimize the objective without any constraints, and fixed-weight linear convex combinations of throughput and fairness do not ensure a prescribed fairness level. To address these conflicting issues, we incorporate the fairness constraint into the maximization process via a Lagrangian approach, namely a hinge-loss penalty. 
\begin{equation}
\label{eq:hinge_loss}
\begin{aligned}
\mathcal{L}
= -\Big( \overline{S} \;+\; \lambda\,\min\!\big(\overline{J}-J_{LB},\,0\big) \Big),\\
\overline{S}
\triangleq \frac{1}{N_b}\sum_{k=1}^{N_b} \tilde{S}_k,
\quad
\overline{J}
\triangleq \frac{1}{N_b}\sum_{k=1}^{N_b} J_k,
\end{aligned}
\end{equation}
which explicitly penalizes violations when $J <J_{LB}$ by a tunable dual variable $\lambda$, thereby preserving controllable fairness while still prioritizing sum-rate improvement.

In the loss calculation \cref{eq:hinge_loss}, we operate on batches of size $N_b$. Let $R_{k,u}$ denote the $u$-th user rate in the $k$-th stream of the batch. The $k$-th sum rate $S_k = \sum_{u=1}^{N_u} R_{k,u}$. Similarly, the $k$-th Jain's index $J_k$ is computed by \cref{jain's index}. 
Because $J_k\in(0,1)$, the sum-rate term must be normalized to a comparable scale; otherwise, the network will over-concentrate on throughput and ignore the fairness. We therefore normalize the sum rate per batch with Max-Min scaling:
\begin{equation}
\label{Eq:Max-Min Normalization}
\tilde S_k=\frac{S_k-S_{\min}}{S_{\max}-S_{\min}},\quad 
\end{equation}
After the scaling, the mean sum rate $\overline{S}$ and mean Jain's index $\overline{J}$ are calculated over the batch. The advantages of Max-Min scaling are that it leverages both extremes, reduces cross-batch fluctuations in the normalization factor, and aligns the normalized sum rate scale with $J\in(0,1)$, which together improve training stability.
In contrast, other methods such as batch-max scaling or Z-score standardization are less suitable: batch-max scaling focuses on the single largest value in each batch, which can induce cross-batch loss fluctuations as $S_{\max}$ varies between batches, while Z-score standardization drives the batch-averaged sum-rate term to zero (although it keeps gradients nonzero). It can harm backpropagation effectiveness when $J >J_{LB}$, if used directly in a combined loss.

Building on the above loss, \Cref{alg:dual training} automatically tunes the dual multiplier \(\lambda\) to enforce target fairness while maximizing the objective in \cref{eq:sumrate_fair_opt}. Here, $\Theta$ denotes the set of trainable parameters of the proposed DNN, including all weights and biases, which maps the input channel realizations to the predicted beamformers. In each iteration, the network updates $\Theta$ and predicts $\hat{\mathbf{f}}_u$. The initial values of \(\lambda\) and \(\Theta\) are insensitive to the final result; the sensitive hyperparameters are the dual step \(\eta_\lambda\) and the tolerance \(\epsilon\). A larger \(\epsilon\) relaxes the constraint and can drift away from the target \(J_{LB}\),  whereas a too small \(\epsilon\) leads to harder convergence because Jain's index fluctuates across each training. Similarly, an overly large dual step size induces update fluctuations of \(\lambda\), while a too small \(\eta_\lambda\) does not have enough power to penalize the sum rate term. Hence, \(\eta_\lambda\) and \(\epsilon\) must be chosen jointly to balance stability and accuracy.
\begin{algorithm}[t]
\begin{algorithmic}[1]
\caption{Adaptive Dual-Multiplier Training for Fairness-Constrained Sum-Rate Maximization}
\label{alg:dual training}
\STATE Initialize $\Theta^{(0)}$, $\lambda^{(0)}>0$, $J_{LB}\!\in\!(0,1)$, $\eta_\lambda$, $\delta$, and $\epsilon$; 
\STATE Set the outer loop counter $n\!\leftarrow\!1$, and the inner loop steps $M\!\leftarrow\!\lceil N_{\text{train}}/N_b\rceil$
\REPEAT
  \FOR{$m=1$ \textbf{to} $M$}
    \STATE Input training batch $\mathbf{H}_{bu}^{(m)}$
    \STATE Predict beamformer $\mathbf{\hat{f}}_u^{(m)}$ with $\Theta^{(m)}$
    \STATE Compute $k$-th $S_k^{(m)}$ and $J_k^{(m)}$; normalize $S_k^{(m)}$ via~\cref{Eq:Max-Min Normalization}
    \STATE Compute $\overline{S}^{(m)}$, $\overline{J}^{(m)}$, and $V^{(m)}\ \gets\overline{J}^{(m)}-J_{LB}$
    \STATE Compute loss $\mathcal{L}^{(m)}$ in ~\cref{eq:hinge_loss}
    \IF{$|V^{(m)}|>\epsilon$}
      \STATE $\lambda^{(m)}\gets\lambda^{(m-1)}+\eta_\lambda\,(J_{LB}-\overline{J}^{(m)})$ 
    \ELSIF{$|V^{(m)}|\leq\epsilon$}
      \STATE $\lambda^{(m)}\gets\lambda^{(m-1)}$
    \ENDIF
    \STATE $\Theta^{(m)}\leftarrow\text{Adam}\big(\Theta^{(m-1)},\,\nabla_\Theta \mathcal{L}^{(m)}\big)$
  \ENDFOR
  \STATE $n\leftarrow n+1$
\UNTIL{$\|\nabla_\Theta \mathcal{L}^{(n)}\|\le\delta$ \textbf{and} $|V|^{(n)}\le\epsilon$}
\RETURN $\Theta^{(n)}$, $\lambda^{(n)}$, $\mathbf{\hat{f}}_u^{(n)}$
\end{algorithmic}
\end{algorithm}
\section{Simulations}
In this section, we present the simulation result to demonstrate the performance of the proposed adaptive fairness control algorithm in \cref{sec:WiT}. We use the wSLNR as defined in \cref{eq:wSLNR,eq:weighted_value,eq:wbeamformer}  to generate the benchmark solutions, and we employ target $J_{LB}$ to train the \ac{DNN}. For comprehensive comparison across scenarios, we also include \ac{MRT}, \ac{ZF}, and conventional SLNR as baseline beamforming schemes. 
\subsection{Simulation Setup}
We simulate a single-cell wireless setup in which a base station (\ac{BS}) equipped with $N_t=16$ antenna elements serves $N_u=12$ single antenna users. The system operates in the sub-6~GHz band with a center frequency of 2~GHz. Users are randomly positioned within a circular area of radius 500~m centered at the \ac{BS}. The total transmit power is \(P_{\text{tot}}=10\,\text{W}\). For each \ac{BS}--\ac{UE} link, we assume a noise-limited Rayleigh fading channel to enable efficient training data generation. This setup is both computationally efficient and theoretically sufficient to realize the fairness–sum-rate Pareto front, because we only target the optimization at a certain specified fairness level. Additionally, it enables us to assess the feasibility of a task-specific WiT architecture, building on our previous work across various tasks \cite{9833994}.

Building on the above setup, we generate \(50\,000\) independent samples by randomly placing the \(N_u\) users within the cell and computing the narrowband \ac{BS}--\ac{UE} channels for each sample. We then partition the dataset into training set, validation set, and test set: \(N_{\text{train}}=32\,000\), \(N_{\text{val}}=8\,000\), and \(N_{\text{test}}=10\,000\). The key training hyperparameters of the \ac{DNN} are summarized in \cref{tab:dnn_hyperparameters}.

We use the fairness achieved by \ac{wSLNR} with different $\alpha_\ell$ as the lower bound $J_{LB}$ for \ac{DNN} training. The DNN’s achieved fairness is denoted $J_{\mathrm{DNN}}$. Key statistics are summarized in~\cref{tab:key_statistics}.
\begin{table}[!t]
\centering
\caption{Hyperparameters of the DNN }
\label{tab:dnn_hyperparameters}
\small
\setlength{\tabcolsep}{3pt}
\renewcommand{\arraystretch}{1.15}
\begin{tabular}{|p{0.32\columnwidth}|p{0.12\columnwidth}|p{0.32\columnwidth}|p{0.12\columnwidth}|}
\hline
\multicolumn{2}{|c|}{\textbf{WiT parameters}} & \multicolumn{2}{c|}{\textbf{Training parameters}} \\
\hline
Emb. Factor ($N_e$)       & 4                & Batch size ($N_b$)         & $2^8$ \\
Blocks ($N_{\text{att}}$) & 8                & Init.\ LR                  & 0.002 \\
Heads ($N_{\text{head}}$) & 4                & Tolerance ($\epsilon$)      & 0.003 \\
Feature dim.\ ($N_f$)     & 33               & Dual step ($\eta_\lambda$)  & 0.01 \\
\hline
\multicolumn{4}{|c|}{\textbf{Total Learnable Parameters.}\ $N_{\text{L}}=8.73\times10^{5}$} \\
\hline
\end{tabular}
\end{table}
\subsection{Performance Evaluation}
\begin{figure}
    \centering
    \includegraphics[width=1\columnwidth]{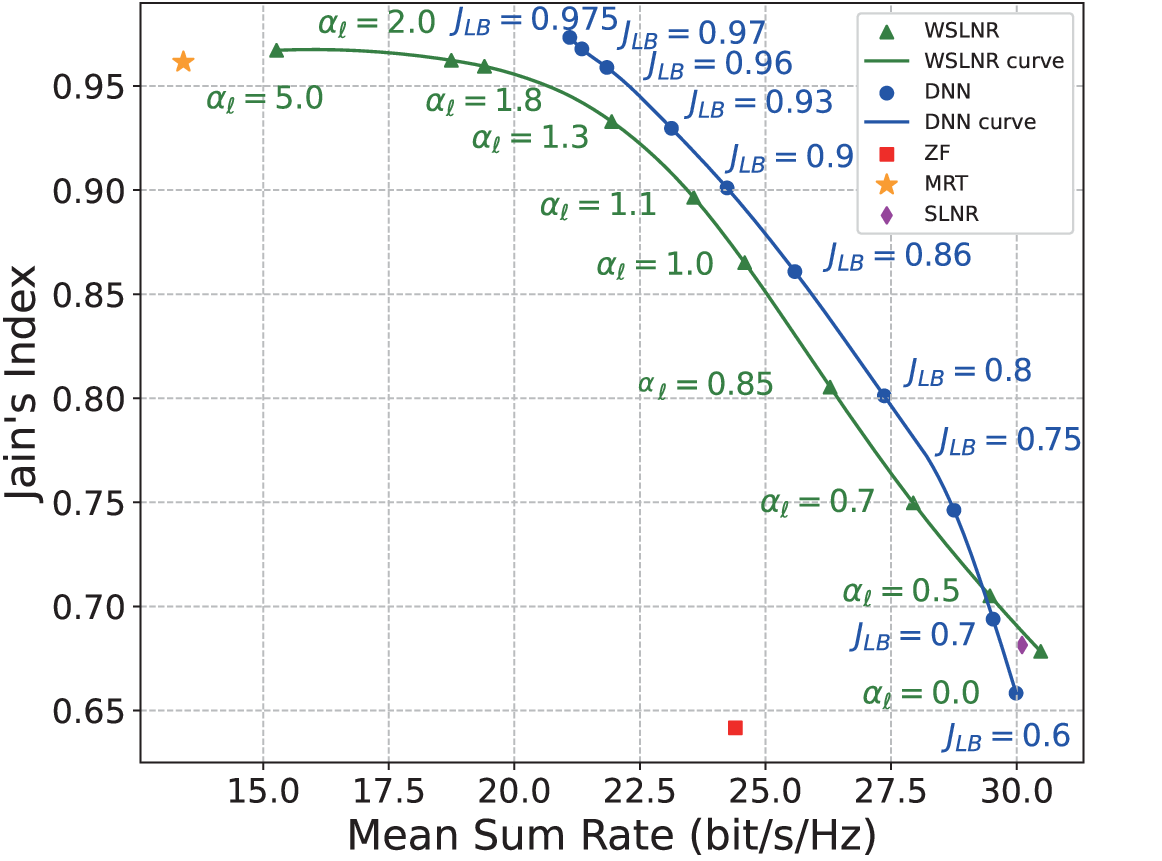}
    \caption{Cross comparison on mean achievable rate and fairness between wSLNR $(\alpha_\ell)$ and DNN trained with target $J_{LB}$.}
    \label{fig:scatter}
\end{figure}

In \cref{fig:scatter}, we obtain 10 points that are used to predict the trace of the Pareto front achievable by the proposed \ac{DNN} approach (blue curve in \cref{fig:scatter}). During training, we observe a special case: choosing $J_{LB}=0.6$ is effectively equivalent to sum-rate maximization, because the Jain's index remains above $0.65$ even when only the sum rate is maximized. Nevertheless, this solution is not globally sum-rate optimal, as \ac{wSLNR} with $\alpha_\ell = 0$ achieves a higher mean sum rate, as shown in \cref{fig:scatter}. Subsequently, the second term in \cref{eq:hinge_loss} is always zero. The two curves intersect near $J=0.7$, beyond which the \ac{DNN} starts to outperform \ac{wSLNR}. As fairness increases, the mean sum rate of \ac{wSLNR} degrades more rapidly than that of the DNN. Consistent with the same issue in ~\cite{9991855,10994230}, \ac{wSLNR} does not optimize the sum rate explicitly at a prescribed fairness level. Instead, fairness is calculated indirectly through $\alpha_\ell$, and the inverse-gain weighting $\omega_\ell$. On the one hand, a larger $\alpha_\ell$ over-targets weak users and over-penalizes strong users; on the other hand, the sum rate is not calculated at a fixed fairness level.
By contrast, the \ac{DNN} is trained with the hinge loss so that the adaptive dual $\lambda$ explicitly anchors the fairness at a certain level within a small tolerance $\epsilon$ and maximizes the overall achievable rate once the fairness constraint is satisfied. Consequently, for larger target fairness, the \ac{DNN} exhibits better performance. For a complete cross-comparison, we list the converged adaptive $\lambda$ and final fairness $J_\mathrm{DNN}$ in \cref{tab:key_statistics}.
\begin{table}[!t]
\centering
\renewcommand{\arraystretch}{1.15}
\setlength{\tabcolsep}{3.4pt}
\caption{Performance comparison: adaptive fairness training vs.\ \ac{wSLNR}, where Mean SR denotes the achievable mean sum rate in bit/s/Hz.}
\label{tab:key_statistics}
\begin{tabular}{|c|c|c|c|c|c|}
\hline
\multicolumn{2}{|c|}{\textbf{wSLNR}} & \multicolumn{4}{c|}{\textbf{DNN}} \\ 
\hline
$\alpha_\ell$ & Mean SR & $J_{LB}$ & $J_{\mathrm{DNN}}$ & $\lambda$ & Mean SR \\
\hline
 0    & 30.48 & 0.60 & 0.658 & 0.000 & 29.99 \\
 0.5  & 29.47 & 0.70 & 0.693 & 0.897 & 29.53 \\
 0.7  & 27.94 & 0.75 & 0.746 & 1.274 & 28.75 \\
 0.85 & 26.29 & 0.80 & 0.801 & 2.200 & 27.36 \\
 1.0  & 24.59 & 0.86 & 0.861 & 2.212 & 25.58 \\
 1.1  & 23.57 & 0.90 & 0.901 & 2.467 & 24.23 \\
 1.3  & 21.94 & 0.93 & 0.929 & 2.677 & 23.12 \\
 1.8  & 19.40 & 0.96 & 0.958 & 3.002 & 21.84 \\
 2.0  & 18.75 & 0.97 & 0.967 & 3.421 & 21.34 \\
 5.0  & 15.26 & 0.973& 0.973 & 3.771 & 21.10 \\
\hline
\end{tabular}
\end{table}

\begin{figure}
    \centering
    \vspace{-1.2em}
    \includegraphics[width=\columnwidth]{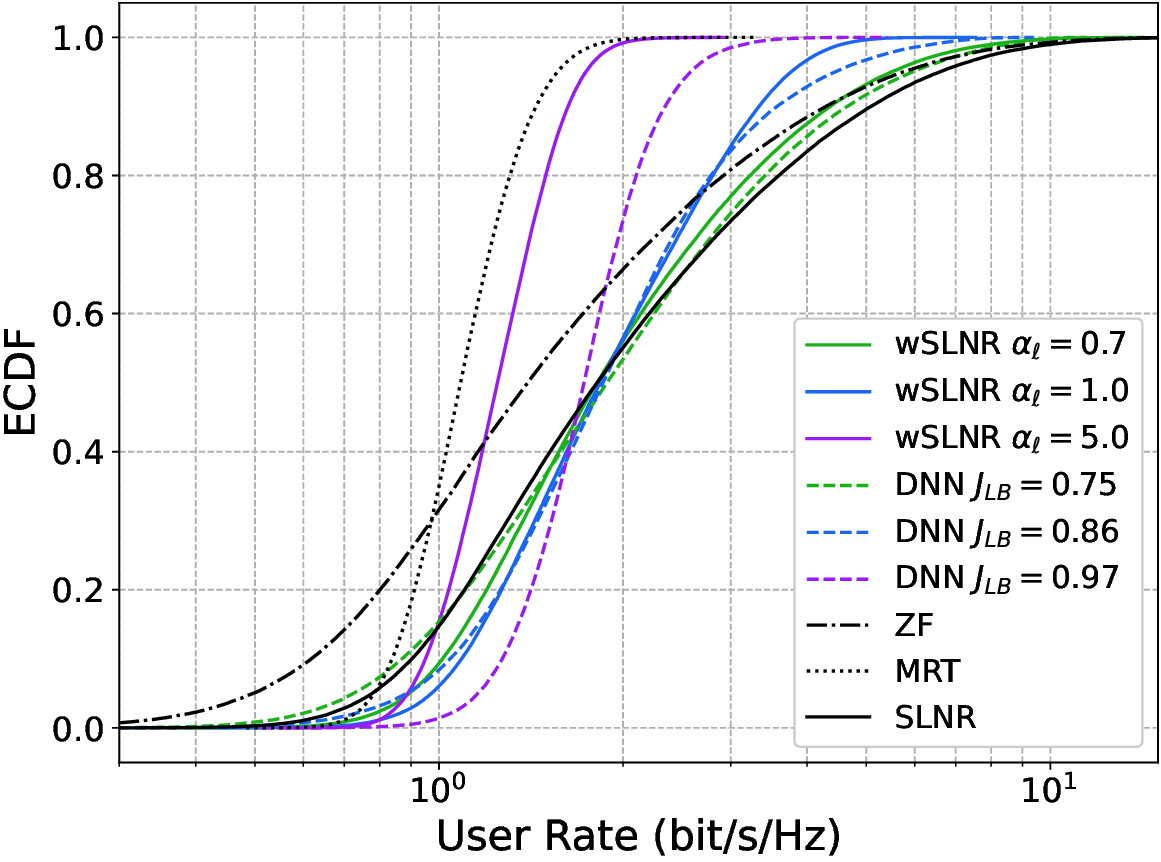}
    \caption{User-rate ECDF at selected target fairness values from \cref{tab:key_statistics}; \ac{DNN} (dashed) vs.\ \ac{wSLNR} (solid).}
    \label{fig:user_cdf}
\end{figure}

\begin{figure}
    \centering
    \includegraphics[width=\columnwidth]{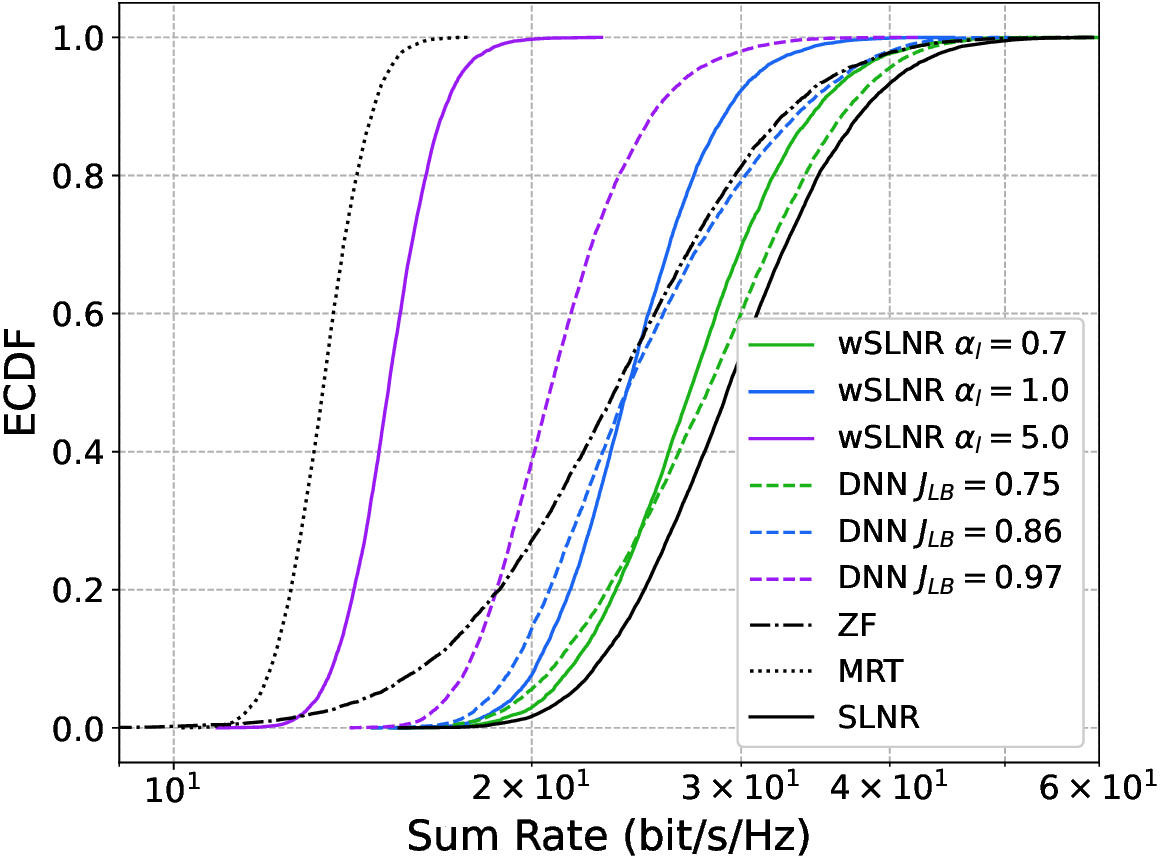}
    \caption{Sum-rate ECDF at selected target fairness values from \cref{tab:key_statistics}; \ac{DNN} (dashed) vs.\ \ac{wSLNR} (solid).}
    \label{fig:sum_cdf}
\end{figure}

Next, we select comparable fairness pairs from \cref{tab:key_statistics} to evaluate both user-rate and sum-rate performance. Specifically, we train DNNs with target fairness levels $J_{LB}\in\{0.75,\,0.86,\,0.97\}$ and use the corresponding \ac{wSLNR} baselines configured with $\alpha_\ell \in \{0.7,\ 1.0,\ 5.0\}$.
As shown in \cref{fig:user_cdf}, for fairness levels below \(0.86\), \ac{wSLNR} and \ac{DNN} exhibit similar performance for weak users (10th percentile), whereas the \ac{DNN} achieves higher rates for strong users (90th percentile) while maintaining weak-user performance. For larger fairness levels (e.g., $J>0.9$), the \ac{DNN} delivers substantial gains across nearly all users. This advantage is amplified in the sum-rate performance shown in \cref{fig:sum_cdf}, where the improvement  for strong users is especially enhanced.

\begin{figure}
    \vspace{-1.6em}
    \centering
    \includegraphics[width=\columnwidth]{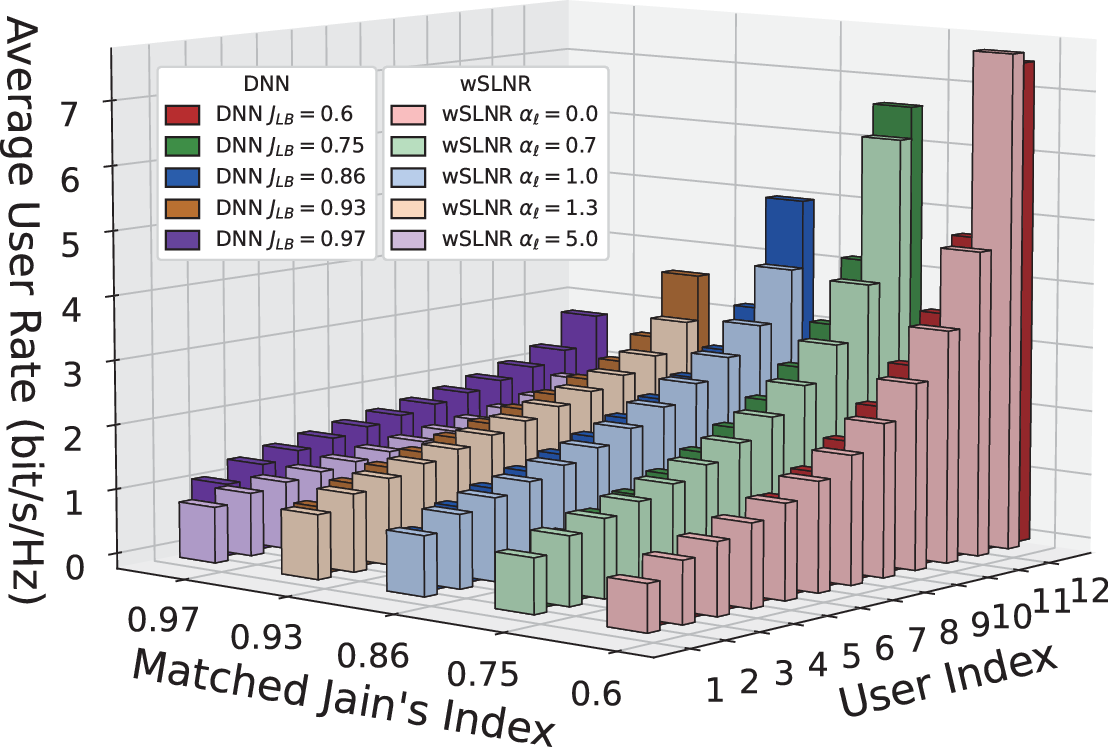}
    \caption{Per-user rate comparison with users ordered by index (weakest to strongest) at selected matched-fairness pairs from \cref{tab:key_statistics}: \ac{DNN} (darker bars) vs.\ \ac{wSLNR} (lighter bars).}
    \label{fig:bars-3d}
\end{figure}

\begin{figure} 
    \centering
    \includegraphics[width=\columnwidth]{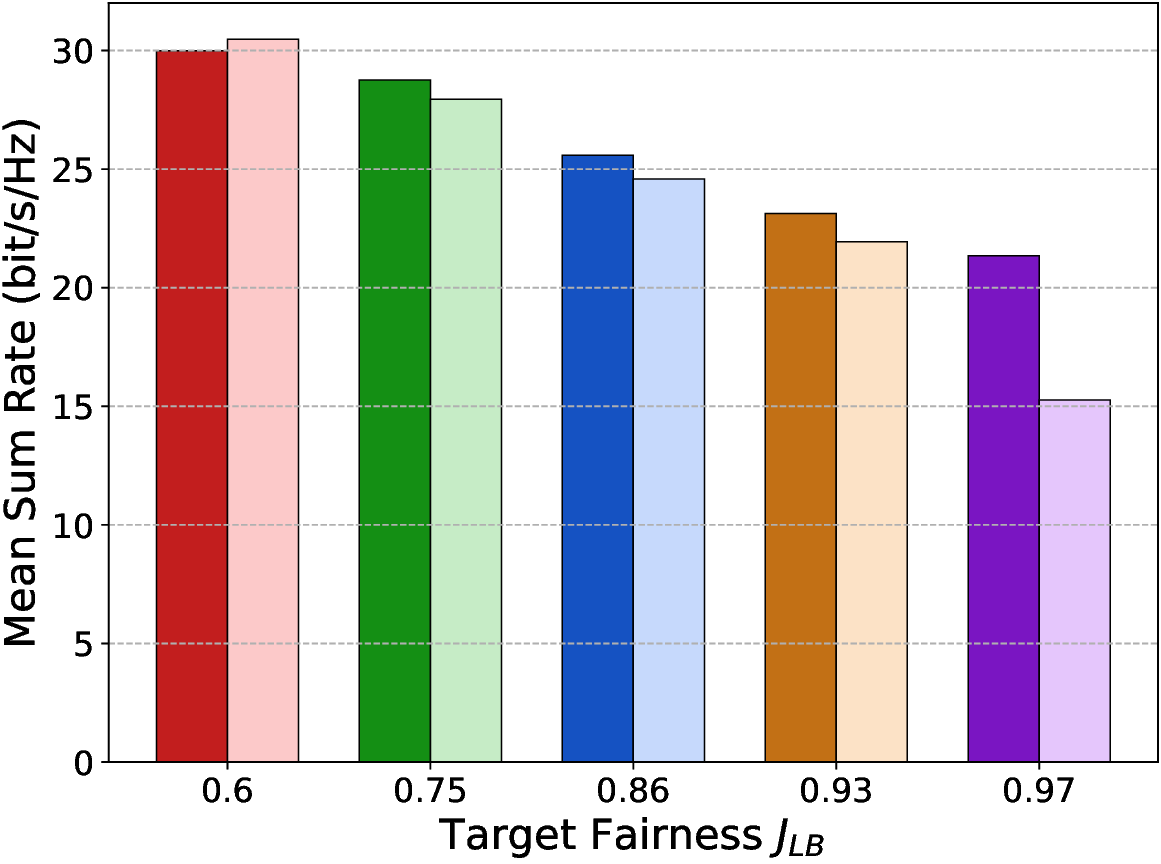}
    \caption{Mean sum rate comparison at selected matched-fairness pairs from \cref{tab:key_statistics}: \ac{DNN} (darker bars) vs.\ \ac{wSLNR} (lighter bars).}
    \label{fig:bars-2d}
\end{figure}

To make the per-user comparison explicit, we sort the user rates in ascending order, align them by user index to produce the bar chart in \cref{fig:bars-3d}. We use five pairs for a comprehensive comparison: \ac{DNN} with $J_{LB}=\{0.6,\,0.75,\,0.86,\,0.93,\,0.97\}$ versus the corresponding \ac{wSLNR} with $\alpha_\ell \in \{0.0,\ 0.7,\ 1,0,\ 1.3,\ 5.0\}$. Aggregating these bars yields the mean sum rate comparison in \cref{fig:bars-2d}, which mirrors the trends observed in \cref{fig:user_cdf} and \cref{fig:sum_cdf}: the larger the fairness target, the better the \ac{DNN} performs.

The underlying mechanism is as follows: The \ac{DNN}, trained with a hinge loss and an adaptive dual $\lambda$, effectively drives the fairness into a certain region (within tolerance $\epsilon$) and then maximizes the achievable rate subject to that constraint. As can be seen from the fifth column in \cref{tab:key_statistics}, higher fairness targets require larger $\lambda$ to counteract the natural tendency of pure sum–rate maximization, which punishes the fairness. Although an exponential hinge loss can further strengthen the penalty, in practice, it over-penalizes throughput at lower fairness targets by over-emphasizing fairness. Thus, $\lambda$ remains the key tuning parameter that directly controls the strength of the sum-rate penalty in the hinge loss.
\section{Conclusion}
In this work, we present an optimization-based learning approach using the \ac{WiT} architecture, which leverages \ac{CSI} as \ac{DNN} input data and addresses an adjustable sum-rate versus fairness trade-off. Specifically, we employ a Lagrangian approach to formulate a hinge-loss objective and adaptively update the dual multiplier $\lambda$ by dual ascent. The resulting curve is the Pareto front of our approach, which may remain strictly suboptimal. Moreover, this approach provides an adaptive control for balancing conflicting objectives and is therefore well-suited for tasks that require enforcing the constraints to a prescribed level. Looking ahead, our future work will investigate whether the \ac{DNN} converges to the optimal solution of the fundamental optimization problem in~\cref{eq:sumrate_fair_opt}. We will also extend the scenario setup and incorporate additional objectives such as energy efficiency.
\paragraph*{Acknoweledgment} This work has been funded by the Vienna Science and Technology Fund (WWTF) [Grant ID: 10.47379/ICT25005].

\bibliographystyle{IEEEtran}
\bibliography{Bibliography}

\end{document}